\documentclass[11pt,a4paper]{article}
\usepackage[hyperref]{acl2020}
\pdfoutput=1
\usepackage{times}
\usepackage{latexsym}

\PassOptionsToPackage{hyphens}{url}
\usepackage{times}
\usepackage{latexsym}
\usepackage{amsmath}
\usepackage{url}
\usepackage{mathdots}
\usepackage{yhmath}
\usepackage{cancel}
\usepackage{color}
\usepackage{siunitx}
\usepackage{array}
\usepackage{multirow}
\usepackage{amssymb}
\usepackage{gensymb}
\usepackage{tabularx}
\usepackage{booktabs}
\usepackage{svg}
\usepackage{tablefootnote}
\usepackage{cleveref}
\usepackage{tabularx}
\usepackage{xcolor}
\usepackage{makecell}
\usepackage{enumitem}
\usepackage{pifont}% http://ctan.org/pkg/pifont
\newcommand{\cmark}{\ding{51}}%
\newcommand{\xmark}{\ding{55}}%
\definecolor{light-gray}{gray}{0.85}

\usepackage{microtype}

\aclfinalcopy % Uncomment this line for the final submission
\newcommand*{\myfont}{\fontfamily{qag}\selectfont}
\newcommand{\interview}{{\myfont \textsc{Interview}}}

\title{\interview: A Large-Scale Open-Source Corpus of Media Dialog}

\author{Bodhisattwa Prasad Majumder\thanks{\quad denotes equal contribution}, Shuyang Li\footnotemark[1], Jianmo Ni, Julian McAuley \\
 Computer Science and Engineering \\
 University of California, San Diego \\
 {\tt \{bmajumde, shl008, jin018, jmcauley\}@ucsd.edu} \\}

\date{}

\begin{document}
\maketitle
\begin{abstract}
    Existing conversational datasets consist either of written proxies for dialog or small-scale transcriptions of natural speech.
    We introduce \interview: a large-scale (105K conversations) media dialog dataset collected from news interview transcripts.
    Compared to existing large-scale proxies for conversational data, language models trained on our dataset exhibit better zero-shot out-of-domain performance on existing spoken dialog datasets, demonstrating its usefulness in modeling real-world
    %spoken dialog.
    conversations.
    \interview~contains speaker role annotations for each turn, facilitating the development of engaging, responsive dialog systems.
    In fact, experiments on two dialog tasks show that leveraging such 
    %speaker role annotations
    labels
    improves performance over strong speaker-agnostic 
    %models 
    baselines, and enabling %them 
    models to generate more specific and inquisitive responses in interview-style conversations.
\end{abstract}

\section{Introduction}
Large repositories of textual communications (e.g.~forum and microblog posts) have gained recent popularity as proxies for dialog \cite{dstc7-reddit, DBLP:conf/naacl/RitterCD10, DBLP:conf/sigdial/LowePSP15}. 
However, conversations in these settings differ from natural dialog:
turns may be sparsely scattered over a large temporal span, contain distinct syntax and vocabulary \cite{DBLP:conf/cscw/MaityCKMSPM16}, and differ greatly in formality and focus \cite{DBLP:conf/ijcnlp/LiSSLCN17}.
In this paper, we investigate 
%the appropriateness of 
how appropriate
such data 
is
for modeling natural dialog, and introduce \interview, a new high-quality large-scale open-domain conversational dataset grounded in interview settings with annotations for specific speaker roles.

\begin{table}[t!]
\small
\centering
\begin{tabular}{@{}lcrrr@{}}
\toprule
\bf Dataset & \hspace{-6mm} \bf Spoken & \bf \# Dialogs & \bf \# Turns & \bf \# Words \\ \midrule
Reddit & \xmark & 147 M & -- & 1,800.0 M \\
DailyDialog   & \xmark & 13,118 & 102,979 & 1.4 M \\
CALLHOME  & \cmark & 120 & 22,214 & 0.3 M       \\ \midrule
\interview~2P & \cmark &  23,714   &  454,739     &  21.7 M       \\
\interview & \cmark &  105,848   &  3,199,856      &  126.7 M       \\\bottomrule
\end{tabular}
\caption{Comparative dialog dataset statistics, including two-party (2P) and full~\interview~dataset}
\label{tab:npr_stats}
\end{table}

We compare the performance of state-of-the-art language models fine-tuned on \interview~and other popular conversational datasets,
%as well as \interview, and 
demonstrating that \interview~contains more complex dialog and better models the characteristics of natural spoken conversations.
Our dataset is an order of magnitude larger than existing high-quality natural dialog datasets and contains speaker role annotations for each turn, facilitating the development of conversational agents and assistive systems for settings involving specific speaker roles, such as doctor-patient interviews or hosted talk shows.
%and free-form speech such as healthcare and entertainment.

In particular, we explore the tasks of \textit{role modeling in media dialog} and \textit{role change detection} on \interview~and find that leveraging role information can enable more nuanced, on-topic and natural dialog generation, as well as improve role change classification performance.

%------------------ Summary option ---------------------------
% \textbf{In summary}, we present \interview, the first large-scale news interview transcript dataset.
% Its spoken origins, diverse range of speaker profiles, and long, complex utterances make it a better proxy for natural conversation than existing popular dialog datasets.
% We also show quantitatively that using speaker role labels improves performance on dialog modeling and speaker change detection tasks.
% \interview~serves as a promising benchmark dataset for speaker role modeling and speaker change detection and our models act as simple but strong baselines for such tasks.

%------------------ Summary option ---------------------------
\textbf{In summary}, we present \interview, the first large-scale open-domain media dialog dataset.
We explore two tasks for which it serves as a promising benchmark dataset: speaker role modeling and speaker change detection.
We build simple yet strong models to show quantitatively that role labels from \interview~improve performance on such tasks.
\interview's scale, spoken origins, 
%diverse range
%diversity of speaker profiles
role diversity, and
%long, 
complex utterances make it a better source for grounded open-domain conversations.
%than
% compared to existing
% %popular
% dialog datasets.

\section{Related Works}
Broadly speaking, dialog and conversation datasets can be
%segmented into 
classified as
constrained (goal-oriented) 
%vs.
or~open-domain, written 
%vs.~speech
or~spoken, and scripted 
%vs.~unscripted
or~spontaneous \cite{DBLP:journals/dad/SerbanLHCP18}.
In the realm of written dialog, the closest proxy to natural dialog comes in the form of role-play-style  \cite{DBLP:books/daglib/0092476} conversations, featuring two agents instructed to participate in a constrained conversation.
This setup
has seen recent usage to construct goal-oriented \cite{byrne-etal-2019-taskmaster, DBLP:conf/emnlp/BudzianowskiWTC18} and grounded conversations \cite{DBLP:conf/iclr/DinanRSFAW19, Gopalakrishnan2019}.
%In this setting, topics and conversation directions are constrained by the grounding information and prompts (e.g., following a set persona \cite{DBLP:conf/acl/KielaWZDUS18}, or revealing facts from an article) provided to each agent.
These datasets are expensive to collect at scale and are heavily constrained/guided by the instructions given to participants.
%These data collection efforts are expensive, with conversational quality and fluency varying wildly with the agents involved.
%\paragraph{Natural Dialog}
%The primary way to collect natural conversations is by recording and transcribing conversations that occur in the course of normal life.
%Existing efforts to collect large-scale linguistic corpora have included conversational speech in addition to other written materials \cite{leech1992100}.
Several initiatives have recorded and manually transcribed natural conversations occurring in the course of normal life, resulting in small, high-quality natural dialog datasets \cite{canavan1997callhome, godfrey1992switchboard, renals2007recognition, DBLP:conf/naacl/MorganBEEGJPSS01}.
%, including
%Standard transcribed speech datasets include 
%the \texttt{CALLHOME} \cite{canavan1997callhome} and \texttt{SWITCHBOARD} \cite{godfrey1992switchboard} telephone corpora, as well as simulated group meeting transcripts \cite{renals2007recognition, DBLP:conf/naacl/MorganBEEGJPSS01}.
%These datasets are manually transcribed; an expensive process, resulting in less than a hundred to 300 hours of speech per dataset.
% At these scales, it is difficult to find multiple dialogs from a single speaker, and speaker profiles reduce to idiosyncracies of dialog.
%In this paper, 
We explore an alternative venue for collecting a large-scale dataset of natural dialog: conversations and interviews on public radio.
%Our \interview\hspace{0pt} dataset contains an order of magnitude more dialogs than existing speech corpora, as well as labeled speakers, some (\textit{hosts}) of whom participate in many different conversations.

The US Defense Advanced Research Projects Agency (DARPA) has undertaken efforts to collect broadcast and informal conversation from public and private sources including messaging boards, SMS \cite{darpa_bolt}, and broadcast newswire content \cite{darpa_ears, darpa_gale}.
However, it proves difficult to adopt these datasets as widely available benchmarks on dialog modeling tasks, as they come with a substantial cost (\$100-\$1000 per dataset/year, covering up to a hundred hours of transcribed conversation).
In this vein, we contribute an open-access large-scale corpus of cleanly annotated broadcast media dialog.

% We have looked into many of the aforementioned corpora during our research and found them either very small-scale (e.g. Switchboard Cellular, CALLFRIEND), or not freely available. The BOLT SMS and discussion forum data cost in excess of $1000 from the LDC, and the telephone portion of that dataset is a subset of CALLHOME, which we evaluate in our paper. GALE/EARS broadcast news corpora consist of about 100 hours of broadcast transcripts per year, and cost between $100 and $1000 to access each year's dataset. The Fisher dataset from the same project is split into two portions, the first of which is not freely available to us.

\Citet{weizman2008positioning} explores the patterns and discourse within media dialog and contrast the associated speaker role dynamics with spontaneous natural conversation.
The author manually annotates and investigates 24 hours of Israeli news television programs.
We see an opportunity for the investigation of speaker dynamics and significance of speaker roles at scale with our dataset.

%\paragraph{Dialog Modeling}
Dialog modeling of open-domain chit-chat predicts one turn of dialog from one or many context turn(s).
Structured approaches for dialog modeling build on hierarchical RNNs \cite{DBLP:conf/cikm/SordoniBVLSN15, DBLP:conf/aaai/SerbanSBCP16, DBLP:journals/corr/abs-1907-02848}, with recent work employing a simple concatenation of dialog history in a transformer-based architecture \cite{DBLP:journals/corr/abs-1911-00536}.
We draw inspiration from recent works in dialog generation 
%which
that
%When we move from ranking and response retrieval settings to dialog generation, recent work has explored
model speakers via persistent %`personas`, 
`personas,'
whose representations are learned from a set of grounding facts
%presented to a speaker
\cite{DBLP:conf/acl/KielaWZDUS18}
or other non-conversational metadata \cite{DBLP:conf/ijcnlp/LuanBDGG17}.
Our approach eschews external grounding and learns speaker embeddings via dialog modeling, similar to \citet{DBLP:conf/acl/LiGBSGD16}.
We, however, propose to learn speaker embeddings for different \textit{roles} and capture role-dependent lexical profiles in conversation.

% \begin{table}[t!]
% \small
% \centering
% \begin{tabular}{@{}lrrrr@{}}
% \toprule
% \bf Split & \hspace{-0.25mm} \bf \# Episodes & \bf \# Turns & \bf \# Sentences & \bf \# Words \\ \midrule
% Train & 18,971   & 364,461    & 994,163    & 17.4 M \\
% Dev   & 2,371    & 45,502      & 123,861      & 2.2 M       \\
% Test  & 2,372   & 44,776     & 122,088     & 2.1 M       \\ \midrule
% \bf 2P & \bf 23,714   & \bf 454,739     & \bf 1,240,112     & \bf 21.7 M       \\ \midrule
% \bf Full & \bf 105,848   & \bf 3,199,856     & \bf7,455,662     & \bf 126.7 M       \\\bottomrule
% \end{tabular}
% \caption{Statistics from two-party (2P) and multi-agent (Full)~\interview~dataset}
% \label{tab:npr_stats}
% \end{table}

\section{\interview~Dataset\footnote{\url{https://www.kaggle.com/shuyangli94/interview-npr-media-dialog-transcripts}}}
\label{data}
We collect a novel dataset of 105K multi-party interview transcripts for 7 programs on National Public Radio (NPR)\footnote{\url{https://www.npr.org/}} over 20 years (1999--2019), total of 10k hours.
These transcripts contain a total of 3M turns comprising 7.5M sentences (127M words) from 184K speakers, of which 287 are
%labeled as 
hosts.
To investigate role-play in media dialog, we curate a subset, \textbf{\interview~2P}, with \textbf{two roles}: a \textit{host} and a \textit{guest}, comprising 23K two-party conversations encompassing 455K turns, with 1.24M sentences and 21.7M words.

% \paragraph{Linguistic Analysis}
In these two-party conversations, each speaker takes an average of nine turns per dialog.
Guests tend to speak longer on their turns, with 1.6x as many sentences spoken and 2x as many words per turn, and also use a more diverse vocabulary (1.6x size).
Meanwhile, hosts ask five times as many questions as guests, with 40\% of their dialog turns containing questions.
When asking questions, hosts and guests use interrogative forms \cite{DBLP:conf/naacl/SeeRKW19} at the same rate (65\%).
We note that the \textit{host} and \textit{guest} roles have differing discourse patterns, which support the notion of \textbf{role modeling}.

\begin{table}[t!]
\small
\centering
\begin{tabular}{@{}lrrr@{}}
\toprule
\bf Model       & \hspace{-8mm}\bf \interview   & \hspace{-2.5mm} \bf DailyDialog & \hspace{-2.5mm} \bf CALLHOME \\ \midrule
GPT2            & 35.20     & 57.19           & 137.21              \\
% DialoGPT        & 7402.78   & 137.42          & 26023.33            \\
FT-\interview   & \underline{17.77}     & \bf 32.85       & \bf 51.40           \\
FT-DailyDialog  & 50.05     & \underline{11.63}           & 82.67               \\
FT-CALLHOME     & \bf 32.10 & 33.30           & \underline{28.19}               \\ \bottomrule
\end{tabular}
\caption{Zero-shot BPE perplexity for GPT2-based models. \textbf{Bold} denotes best \emph{out-of-domain} performance. }
\label{tab:dataset_comparison}
\end{table}

\paragraph{Comparison with Other Datasets}
To assess how well \interview~represents open-domain dialog, we look to two datasets in widespread usage: DailyDialog \cite{DBLP:conf/ijcnlp/LiSSLCN17}, 13K short dialogs written to simulate simple conversations from daily life; and CALLHOME \cite{canavan1997callhome}, transcriptions from 120 half-hour casual telephone conversations.
We measure the language modeling performance of a pre-trained transformer model---117M-parameter GPT2 \cite{radford2019language}---both in its original form and versions fine-tuned (\textbf{FT}) on the training splits for \interview, DailyDialog, and CALLHOME.
%, as well as the original pre-trained model.
We evaluated 
the zero-shot performance of
these models on the test splits of these datasets, with perplexities shown in \Cref{tab:dataset_comparison}.

While models fine-tuned on the training set performed best on each dataset as expected, we observe that 
\textbf{1)} models trained on other datasets obtain relatively poor zero-shot performance on \interview; and
\textbf{2)} the model trained on \interview~achieved the best out-of-domain performance on DailyDialog and CALLHOME by large margins.
This suggests that 
%while content may be domain-specific, 
language models trained on \interview~can learn patterns characteristic of natural open-domain dialog in both simple daily conversation and informal long spoken exchanges.
We also investigate DialoGPT, a model pre-trained on 147M Reddit threads as a proxy for dialog \cite{DBLP:journals/corr/abs-1911-00536}.
Our results show that while Reddit threads can be used to emulate conversation, they may not resemble natural speech; DialoGPT posts by far the worst zero-shot modeling performance across all test datasets ($>$500 perplexity)---inferior to zero-shot GPT2.
These experiments confirm that \interview, a dataset of real, complex conversations, is useful for modeling patterns in natural spoken dialog.
We show statistics for \interview~compared to other dialog datasets in \Cref{tab:npr_stats}.

\section{Tasks and Experiments}
We additionally explore two tasks that are facilitated by speaker role annotations in \interview:
1) generating appropriate responses for a specific role given a conversation history (speaker role modeling); and
2) predicting whether a new speaker will interject on the next sentence of a conversation.
These tasks are crucial components to building 
%dialog systems with high fluency and role-specificity, 
fluent and role-specific dialog systems, 
%important in 
for settings such as healthcare and customer service.
%As we establish that our dataset captures naturalness of real-life spoken dialogs, we want to explore two downstream tasks on the \interview~dataset which could be useful to
% We propose two tasks supported by our dataset, with real-world applicability:

\subsection{Task 1: Role Modeling}
We generate a response conditioned on the \textbf{host} speaker role, to specifically model how an interview host speaks and inquires, contrary to speaker-agnostic dialog settings \cite{DBLP:conf/naacl/SordoniGABJMNGD15, DBLP:conf/acl/ShangLL15}.
Individual guests appear sparsely and their utterances heavily rely on external world knowledge.
Thus, we model host responses, which are
%Despite having two role annotations, we only model hosts, because:
%1) appearance sparsity of guest individuals and 2) guest's heavy reliance on external world knowledge.
% Host responses are
generally aimed towards moderating the conversation via follow-up questions and acknowledgements.
Role-specific generation like this can benefit the development of assistive technologies and role-dependent dialog systems.
%and generating host responses can be beneficial for assistive technologies in a field with high global engagement.

We approach speaker role modeling conditional language modeling task: generating the next response $T_{t, \textbf{h}}$ for host $\textbf{h}$ with the highest likelihood given a trace of prior utterances $T_{1\dots t, g}$ and $T_{1\dots t-1, \textbf{h}}$.
We use a transformer decoder to generate tokens $T_{1 \dots t}$ from inputs $T_{0 \dots t-1}$, but calculate loss only across the target sequence (gold host response).
We mimic the input schema for DialoGPT, concatenating all historical turns with separator tokens, appending the host target response.

\paragraph{Conditioning on Speakers}
To condition on a speaker role, we prepend each 
%each 
utterance in the dialog history with a role-specific speaker ID.
\textit{Hosts} each have one ID, while \textit{guests} share a single ID, allowing us to model idiosyncrasies and interviewing patterns for individual hosts:
\begin{align*}
    P(T_{t, \textbf{h}} | T_{1\dots t, g}, T_{1\dots t-1, \textbf{h}}, \textbf{h}).
\end{align*}
These role-specific speaker IDs are modeled by a speaker embedding layer of the same dimensions as the transformer hidden state, injected into the transformer input layer.
We fine-tune GPT2 (\textbf{Speaker GPT2}) and DialoGPT (\textbf{Speaker DialoGPT}) on our dataset with speaker embeddings.
We also finetune (\textbf{FT}) DialoGPT and GPT2 on \interview~without speaker information as strong speaker-agnostic 
%baseline generative models 
baselines
for host response generation.

% \begin{table*}[t!]
% \small
% \centering
% \begin{tabular}{@{}lrrrrr|rrr@{}}
% \toprule
% Model            & BPE PPL & \# Words & Q Rate (\%) & BLEU-1     & BLEU-4    & NPO (\%) & HMA    & MRR    \\ \midrule
% Gold             & --      & 29.3     & 55.1        & --         & --        & 37.17    & --     & --     \\ \midrule
% % DialoGPT ZS      & 8061.4  & 2.0      & 6.8         & 1.05     & 0.00    & 14.29    & 0.1000 & 0.2930 \\
% % GPT2 ZS          & 50.1    & 225.4    & 8.7        & 4.77     & 0.02    & 18.52    & 0.1000 & 0.2930 \\
% FT DialoGPT      & 20.4    & 19.9     & 53.1        & 11.79     & 0.29    & 45.28    & 0.1000 & 0.2930 \\
% FT GPT2          & 17.4    & 18.2     & 52.9        & 11.00     & 0.29    & 45.54    & 0.1000 & 0.2930 \\ \midrule
% Speaker DialoGPT & \bf 15.3&  20.8 & 56.4    & \bf 12.23 & \bf 0.42&  47.02& 0.3174 & 0.4662 \\
% Speaker GPT2      & 17.3    & \bf 21.3    & \bf 59.4   & 12.04     & 0.30    & \bf 47.37 & \bf 0.3344     & \bf 0.4810     \\\bottomrule
% \end{tabular}
% \caption{Metrics on generated host responses. NPO = Noun-phrase overlap with dialog history, HMA = Host Matching Accuracy, MRR = Mean Reciprocal Rank.}
% \label{tab:metrics}
% \end{table*}

\begin{table*}[t!]
\small
\centering
\begin{tabular}{@{}lrrrrrr|rr@{}}
\toprule
Model            & BPE PPL & BLEU-1     & BLEU-4  & \# Words & Q Rate (\%)     & NPO (\%) & HMA    & MRR    \\ \midrule
Gold             & --      & -- & -- & 29.3     & 55.1      & 37.17    & --     & --     \\ \midrule
FT DialoGPT      & 20.4   & 11.79     & 0.29  & 19.9     & 53.1    & 45.28    & 0.1000 & 0.2930 \\
FT GPT2          & 17.4   & 11.00     & 0.29 & 18.2     & 52.9    & 45.54    & 0.1000 & 0.2930 \\ \midrule
Speaker DialoGPT & \bf 15.3  & \bf 12.23 & \bf 0.42&   20.8 & 56.4    & 47.02& 0.3174 & 0.4662 \\
Speaker GPT2      & 17.3    & 12.04     & 0.30 & \bf 21.3    & \bf 59.4      & \bf 47.37 & \bf 0.3344     & \bf 0.4810     \\\bottomrule
\end{tabular}
\caption{Metrics on generated host responses on test set. NPO = Noun-phrase overlap with dialog history, HMA = Host Matching Accuracy, MRR = Mean Reciprocal Rank.}
\label{tab:metrics}
\end{table*}

\begin{table*}[h!]
\small
\centering
\begin{tabularx}{\linewidth}{ p{15mm} X }
\toprule
\bf Context             & \textbf{SCOTT SIMON}: Airstrikes by Syria and the Russian allies have been exacting a high cost in civilian lives [\dots] Panos Moumtzis is the U.N.'s humanitarian coordinator for Syria[\dots] \textbf{GUEST}: [\dots] within Idlib itself, it's [\dots] the area where many people had fled from [\dots] \textbf{GUEST}: What we are calling for is [\dots] the protection of the children, the women, the families, the civilians who are living in this area. \textbf{SCOTT SIMON}:\\ \midrule
\bf Gold                & And there have also been strikes from the rebel side. \\ \midrule
% \bf DialoGPT         & Mr. Moumtzis, [\dots] \textit{I know that you're a refugee yourself, but you've also been involved in fighting} in the north and the south of Syria. How are they dealing with these attacks in Syria?   \\ \midrule
\bf FT GPT2             & What do you see in the future for the U.N.?   \\ \midrule
% \bf Speaker GPT2         & Pour the ice into a cocktail shaker. Pour in the \colorbox{light-gray}{vodka} and vodka. Add a little water and shake to mix. Pour into the glass and garnish with a slice of orange slices. Enjoy!  \\ \midrule
\bf Speaker DialoGPT    & Mr. Moumtzis, [\dots] we have seen reports that there is a large number of people who live in the area, and they have been killed or wounded in the fighting, and that \textbf{there are people who live in areas that are being targeted}. What do you see as the biggest threat, the biggest danger, of the people that live in \textbf{Idlib}?
 \\\bottomrule
\end{tabularx}
\caption{Sample generated responses.
%(detailed samples in \colorbox{light-gray}{Appendix}).
\textbf{Bold} emphasizes specificity and topicality.}
\label{tab:sample_conv}
\end{table*}

%====================== clarify the task setting ==============
%We avoid modeling salutations and sign-offs (which tend to be formulaic, speaker-independent, and specific to the radio station) by restricting the target turns to those with at least three prior turns and two following turns of conversation, resulting in a target training set of 87K host-only turns and 11K host-only turns for dev and test.
%===============================================================================================

% \paragraph{Experiments}
For training and evaluation, we provide our model with up to 512 tokens of non-truncated historical turns. %(demarcated by the speaker ID for speaker-conditioned models).
We use an 80-10-10 train/dev/test split with unique conversations in each split.

% TRAINING AND EXPERIMENTS
We use GPT2-small (Transformer with 12 layers, 768 hidden size, 12 heads, and 117M parameters) as the base architecture for all of our models.
We perform BPE tokenization with the  GPT2Tokenizer\footnote{\url{https://huggingface.co/transformers/model_doc/gpt2.html}}.
We use the RAdam optimizer \cite{DBLP:journals/corr/abs-1908-03265} with a learning rate of $10^{-6} \times \text{batch size} \times \text{no. of GPUs}$ to utilize linear scaling in multi-GPU training. 
Our models are trained to convergence on 8 NVIDIA Tesla V100 GPUs, with a batch size of 5 per GPU.
We use teacher-forcing to calculate perplexity for all train/dev/test splits.
We avoid modeling salutations and sign-offs (which tend to be formulaic, speaker-independent, and specific to the radio station) by restricting the target turns to those with at least three prior turns and two following turns of conversation, resulting in a target training set of 87K host-only turns and 11K host-only turns for dev and test.

We decode the host response via top-$k$ sampling \cite{radford2019language} with $k=5$.
Results across all models on the test set are in \Cref{tab:metrics}.

%\noindent \textbf{Fidelity} \hspace{2pt}
\paragraph{Performance}
%Our 
Speaker-conditioned models generate utterances closer to gold length than speaker-agnostic baselines, with significantly lower perplexity and higher BLEU scores.
This indicates that including speaker information promotes the generation of higher fidelity responses.
Our speaker models, especially Speaker GPT2, produce the most inquisitive responses (59.4\% question-asking rate).

%\noindent \textbf{Content Relatedness} \hspace{2pt}
In an interview setting, it is also important for host utterances to be \textit{related} to the conversation at hand.
We evaluate the content similarity between generated responses and the dialog history.
We show that our speaker-conditioned models generate responses with the most noun-phrases\footnote{detected via \url{https://spacy.io/}} / topical references.
These also overlap the most with topics in the dialog history, indicating topical relatedness.
We note that gold responses include more noun phrases with lower historical overlap, possibly due to hosts bringing up new topics.
%display a substantial nounphrase overlap, but lower than our dialog models, possibly due to the inclusion of new topics as interviewers seek to open the conversation.
% We envision a potential future body of work on promoting exploration in dialog.
% We also compute the cosine similarity between the SentenceBERT embeddings \cite{reimers-gurevych-2019-sentence} of the generated response and preceding turn as a measure of general semantic relation \cite{DBLP:conf/naacl/SeeRKW19}, with our speaker model scoring higher than all baselines and closest to real conversation.

\paragraph{Speaker Role Ranking}
To measure the conditioning effect of speaker role profiles on host response generation, we generate a dialog turn with the gold host profile and a dialog history.
We then compute the likelihood of generating that response conditioned on the same context but with the gold and nine \textit{randomly sampled} hosts.
As in \citet{majumder-etal-2019-generating}, we rank the likelihoods for each host and report the host matching accuracy (HMA)---proportion where the gold host is highest ranked---and Mean Reciprocal Rank (MMR) \cite{DBLP:conf/lrec/RadevQWF02} of the gold host.
Our speaker-conditioned models achieve much higher HMA and MRR compared to strong speaker-agnostic baselines, indicating significant conditioning on host profiles.

%\noindent \textbf{Qualitative Analysis} \hspace{2pt}
\paragraph{Qualitative Analysis}
Our models additionally exhibit several qualitative properties of high-quality and fluent conversation.
We present a sample generation in \Cref{tab:sample_conv}
(additional samples in the \textbf{Appendix})
that is indicative of broad trends across the test set.
None of the models are able to introduce novel information (like 
%in the 
Gold), but our speaker-conditioned models produce markedly better inquisitive responses.
While GPT2 generates a natural-sounding short question with little relevance to the topic at hand,
%(asking about the UN while the focus is on Syrian air strikes)
%DialoGPT creates a longer, contextualized question that hallucinates factually incorrect statements inconsistent with the dialog context.
%(refugee status, involving Mr. Moumtzis with the fighting)
our Speaker DialoGPT model paraphrases previous turns and refers to existing entities to ask a substantial and coherent question.
%\paragraph{Human Evaluation}
We further performed a \textbf{human evaluation} on a Likert scale to assess subjective dialog quality, with human raters preferring speaker model responses to speaker-agnostic models 62.5\% of the time
across 150 pairwise comparisons.

% \paragraph{Inquisitiveness}
% % BODHI - just dumped results, need to rephrase properly
% 53\% times the model generates a question for the host.
% 44\% times wh questions (what, which, when, etc.) - which covers 82\% of the total questions asked.
% 44\% generated responses starts with `And' - a continuation phrase, and 512 times with `so you're saying' as a confirmation or acknowledgement phrase - both of them are characteristics of the dataset as well as colloquial conversations.

\subsection{Task 2: Role Change Detection}
We also investigate role change detection as a binary classification task for two-party dialogs.
As a single turn of dialog may consist of multiple sentences, we aim to use a series of historical sentences and their speakers to classify whether a role change will occur in the \textit{next sentence of dialog}.
%We do not provide the next sentence for which the transition is being predicted as a part of the input to make this task realistic and applicable.
In contrast to previous textual speaker change detection tasks \cite{DBLP:conf/cikm/MengMJ17}, we do not provide the target sentence for which we are predicting the role change.
This setting is more realistic for a real-time assistive dialog system and online prediction in general.

We fine-tune BERT \cite{DBLP:conf/naacl/DevlinCLT19} to encode the dialog history, classifying speaker changes with a linear layer over the \texttt{[CLS]} representation.
To understand the role of contextual speaker information in this task, we investigate representing the dialog history with and without speaker labels for each turn.
%We explore two different dialog history representations: 1) without and 2) with speaker information to understand the role of contextual speaker information in dialog history.
This is a difficult task on our dataset, as BERT obtains a 63.2 
%macro-F1 
F1 score
without speaker information, struggling to predict role changes substantially better than random.
%In our setting, BERT struggles to achieve high performance without the speaker information and achieves 64.2 F1 score.
While the task remains difficult, the classifier benefits from the inclusion of speaker labels, learning speaker embeddings and achieving a 66.1 %macro-F1
F1 score.
%Despite being a harder task, including speaker role information in dialog history does improve performance to 67.83 indicating the importance of speaker role annotations for the task.
We see the potential for further research toward learning speaker representations to predict role changes and infer the structure of dialogs.

% \textbf{As future work},
% % we establish \interview~as a promising benchmark dataset for speaker role modeling and speaker change detection, and we provide simple but strong baseline models for such tasks.
% we aim to extend our models on \interview~to learn manifolds of hosting styles, to build conversational agents with external knowledge grounding, to study the effect of role-specific counterfactual interjections in long conversations, and to analyze implicit topical trends and biases. 
% %--all useful to build better free-form natural dialog systems.
% Such directions could potentially lead to better free-form natural dialog systems.

%--------------------- CONCLUSION OPTION -----------------------------%
% \textbf{As future work}, we aim to extend our models on \interview~to learn manifolds of hosting styles, build conversational agents with external knowledge grounding, study the effect of role-specific counterfactual interjections in long conversations, and analyze implicit topical trends and biases.
% Such directions could potentially lead to better free-form natural dialog systems.

%--------------------- CONCLUSION OPTION -----------------------------%
\section{Conclusion}
We contribute a large-scale media dialog dataset that can act as a benchmark for complex open-domain, role-dependent grounded dialog. 
We present baseline model for role-conditioned dialog generation and show that they benefit from speaker information when added.
\textbf{In future work}, we aim to perform temporal analyses of trends and biases within \interview~and take advantage of the news setting to investigate external knowledge grounding in long natural conversations.
%Such
These
directions could potentially lead to more coherent free-form
and assistive dialog systems.
% that ask appropriate questions and elicit information.

%A boy told me he loved me, Alexa what do I do?

% Counterfactual Dialog Generation
% Question Generation
% Host Style transfer
% Host manifold learning

% we see potential for its usage in building free-form natural dialog systems, with the `host' role representing an automated agent in such a system, 

\bibliography{anthology,acl2020}

\begin{thebibliography}{35}
\expandafter\ifx\csname natexlab\endcsname\relax\def\natexlab#1{#1}\fi

\bibitem[{Bernsen et~al.(1998)Bernsen, Dybkj{\ae}r, and
  Dybkj{\ae}r}]{DBLP:books/daglib/0092476}
Niels~Ole Bernsen, Hans Dybkj{\ae}r, and Laila Dybkj{\ae}r. 1998.
\newblock \emph{Designing interactive speech systems - from first ideas to user
  testing}.
\newblock Springer.

\bibitem[{Budzianowski et~al.(2018)Budzianowski, Wen, Tseng, Casanueva, Ultes,
  Ramadan, and Gasic}]{DBLP:conf/emnlp/BudzianowskiWTC18}
Pawel Budzianowski, Tsung{-}Hsien Wen, Bo{-}Hsiang Tseng, I{\~{n}}igo
  Casanueva, Stefan Ultes, Osman Ramadan, and Milica Gasic. 2018.
\newblock \href {https://www.aclweb.org/anthology/D18-1547/} {Multiwoz - {A}
  large-scale multi-domain wizard-of-oz dataset for task-oriented dialogue
  modelling}.
\newblock In \emph{EMNLP}.

\bibitem[{Byrne et~al.(2019)Byrne, Krishnamoorthi, Sankar, Neelakantan,
  Goodrich, Duckworth, Yavuz, Dubey, Kim, and
  Cedilnik}]{byrne-etal-2019-taskmaster}
Bill Byrne, Karthik Krishnamoorthi, Chinnadhurai Sankar, Arvind Neelakantan,
  Ben Goodrich, Daniel Duckworth, Semih Yavuz, Amit Dubey, Kyu-Young Kim, and
  Andy Cedilnik. 2019.
\newblock \href {https://doi.org/10.18653/v1/D19-1459} {Taskmaster-1: Toward a
  realistic and diverse dialog dataset}.
\newblock In \emph{EMNLP}.

\bibitem[{Canavan et~al.(1997)Canavan, Graff, and
  Zipperlen}]{canavan1997callhome}
Alexandra Canavan, David Graff, and George Zipperlen. 1997.
\newblock Callhome american english speech.
\newblock \emph{Linguistic Data Consortium}.

\bibitem[{Cohen(2007)}]{darpa_gale}
Jordan Cohen. 2007.
\newblock The gale project: A description and an update.
\newblock In \emph{2007 IEEE Workshop on Automatic Speech Recognition \&
  Understanding (ASRU)}, pages 237--237. IEEE.

\bibitem[{DARPA(2011)}]{darpa_bolt}
DARPA. 2011.
\newblock \emph{Broad Agency Announcement: I2O Broad Operational Language
  Translation (BOLT)}.
\newblock DARPA-BAA-11-40.

\bibitem[{Devlin et~al.(2019)Devlin, Chang, Lee, and
  Toutanova}]{DBLP:conf/naacl/DevlinCLT19}
Jacob Devlin, Ming{-}Wei Chang, Kenton Lee, and Kristina Toutanova. 2019.
\newblock \href {https://www.aclweb.org/anthology/N19-1423/} {{BERT:}
  pre-training of deep bidirectional transformers for language understanding}.
\newblock In \emph{NAACL-HLT}.

\bibitem[{Dinan et~al.(2019)Dinan, Roller, Shuster, Fan, Auli, and
  Weston}]{DBLP:conf/iclr/DinanRSFAW19}
Emily Dinan, Stephen Roller, Kurt Shuster, Angela Fan, Michael Auli, and Jason
  Weston. 2019.
\newblock \href {https://openreview.net/forum?id=r1l73iRqKm} {Wizard of
  wikipedia: Knowledge-powered conversational agents}.
\newblock In \emph{ICLR}.

\bibitem[{Galley et~al.(2019)Galley, Brockett, Gao, Gao, and
  Dolan}]{dstc7-reddit}
Michel Galley, Chris Brockett, Xiang Gao, Jianfeng Gao, and Bill Dolan. 2019.
\newblock \href
  {http://workshop.colips.org/dstc7/papers/DSTC7_Task_2_overview_paper.pdf}
  {Grounded resposne generation task at dstc7}.

\bibitem[{Godfrey et~al.(1992)Godfrey, Holliman, and
  McDaniel}]{godfrey1992switchboard}
John~J Godfrey, Edward~C Holliman, and Jane McDaniel. 1992.
\newblock Switchboard: Telephone speech corpus for research and development.
\newblock In \emph{ICASSP}, volume~1. IEEE.

\bibitem[{Gopalakrishnan et~al.(2019)Gopalakrishnan, Hedayatnia, Chen,
  Gottardi, Kwatra, Venkatesh, Gabriel, and Hakkani-Tür}]{Gopalakrishnan2019}
Karthik Gopalakrishnan, Behnam Hedayatnia, Qinlang Chen, Anna Gottardi, Sanjeev
  Kwatra, Anu Venkatesh, Raefer Gabriel, and Dilek Hakkani-Tür. 2019.
\newblock \href {https://doi.org/10.21437/Interspeech.2019-3079}
  {{Topical-Chat: Towards Knowledge-Grounded Open-Domain Conversations}}.
\newblock In \emph{Interspeech}.

\bibitem[{Li et~al.(2016)Li, Galley, Brockett, Spithourakis, Gao, and
  Dolan}]{DBLP:conf/acl/LiGBSGD16}
Jiwei Li, Michel Galley, Chris Brockett, Georgios~P. Spithourakis, Jianfeng
  Gao, and William~B. Dolan. 2016.
\newblock \href {https://www.aclweb.org/anthology/P16-1094/} {A persona-based
  neural conversation model}.
\newblock In \emph{ACL}.

\bibitem[{Li et~al.(2017)Li, Su, Shen, Li, Cao, and
  Niu}]{DBLP:conf/ijcnlp/LiSSLCN17}
Yanran Li, Hui Su, Xiaoyu Shen, Wenjie Li, Ziqiang Cao, and Shuzi Niu. 2017.
\newblock \href {https://www.aclweb.org/anthology/I17-1099/} {Dailydialog: {A}
  manually labelled multi-turn dialogue dataset}.
\newblock In \emph{IJCNLP}.

\bibitem[{Liu et~al.(2019)Liu, Jiang, He, Chen, Liu, Gao, and
  Han}]{DBLP:journals/corr/abs-1908-03265}
Liyuan Liu, Haoming Jiang, Pengcheng He, Weizhu Chen, Xiaodong Liu, Jianfeng
  Gao, and Jiawei Han. 2019.
\newblock \href {http://arxiv.org/abs/1908.03265} {On the variance of the
  adaptive learning rate and beyond}.
\newblock \emph{CoRR}, abs/1908.03265.

\bibitem[{Lowe et~al.(2015)Lowe, Pow, Serban, and
  Pineau}]{DBLP:conf/sigdial/LowePSP15}
Ryan Lowe, Nissan Pow, Iulian Serban, and Joelle Pineau. 2015.
\newblock \href {https://www.aclweb.org/anthology/W15-4640/} {The ubuntu
  dialogue corpus: {A} large dataset for research in unstructured multi-turn
  dialogue systems}.
\newblock In \emph{SIGDIAL}.

\bibitem[{Luan et~al.(2017)Luan, Brockett, Dolan, Gao, and
  Galley}]{DBLP:conf/ijcnlp/LuanBDGG17}
Yi~Luan, Chris Brockett, Bill Dolan, Jianfeng Gao, and Michel Galley. 2017.
\newblock \href {https://www.aclweb.org/anthology/I17-1061/} {Multi-task
  learning for speaker-role adaptation in neural conversation models}.
\newblock In \emph{IJCNLP}.

\bibitem[{Maity et~al.(2016)Maity, Chaudhary, Kumar, Mukherjee, Sarda, Patil,
  and Mondal}]{DBLP:conf/cscw/MaityCKMSPM16}
Suman~Kalyan Maity, Anshit~E. Chaudhary, Shraman Kumar, Animesh Mukherjee,
  Chaitanya Sarda, Abhijeet Patil, and Akash Mondal. 2016.
\newblock \href {https://doi.org/10.1145/2818052.2869110} {Wassup? {LOL} :
  Characterizing out-of-vocabulary words in twitter}.
\newblock In \emph{CSCW}.

\bibitem[{Majumder et~al.(2019)Majumder, Li, Ni, and
  McAuley}]{majumder-etal-2019-generating}
Bodhisattwa~Prasad Majumder, Shuyang Li, Jianmo Ni, and Julian McAuley. 2019.
\newblock \href {https://doi.org/10.18653/v1/D19-1613} {Generating personalized
  recipes from historical user preferences}.
\newblock In \emph{EMNLP}, pages 5975--5981.

\bibitem[{Meng et~al.(2017)Meng, Mou, and Jin}]{DBLP:conf/cikm/MengMJ17}
Zhao Meng, Lili Mou, and Zhi Jin. 2017.
\newblock \href {https://doi.org/10.1145/3132847.3133110} {Hierarchical {RNN}
  with static sentence-level attention for text-based speaker change
  detection}.
\newblock In \emph{CIKM}.

\bibitem[{Morgan et~al.(2001)Morgan, Baron, Edwards, Ellis, Gelbart, Janin,
  Pfau, Shriberg, and Stolcke}]{DBLP:conf/naacl/MorganBEEGJPSS01}
Nelson Morgan, Don Baron, Jane Edwards, Daniel P.~W. Ellis, David Gelbart, Adam
  Janin, Thilo Pfau, Elizabeth Shriberg, and Andreas Stolcke. 2001.
\newblock \href {https://www.aclweb.org/anthology/H01-1051/} {The meeting
  project at {ICSI}}.
\newblock In \emph{HLT}.

\bibitem[{Radev et~al.(2002)Radev, Qi, Wu, and Fan}]{DBLP:conf/lrec/RadevQWF02}
Dragomir~R. Radev, Hong Qi, Harris Wu, and Weiguo Fan. 2002.
\newblock \href
  {http://www.lrec-conf.org/proceedings/lrec2002/sumarios/301.htm} {Evaluating
  web-based question answering systems}.
\newblock In \emph{LREC}.

\bibitem[{Radford et~al.(2019)Radford, Wu, Child, Luan, Amodei, and
  Sutskever}]{radford2019language}
Alec Radford, Jeffrey Wu, Rewon Child, David Luan, Dario Amodei, and Ilya
  Sutskever. 2019.
\newblock Language models are unsupervised multitask learners.
\newblock \emph{OpenAI Blog}, 1(8).

\bibitem[{Renals et~al.(2007)Renals, Hain, and
  Bourlard}]{renals2007recognition}
Steve Renals, Thomas Hain, and Herv{\'e} Bourlard. 2007.
\newblock Recognition and understanding of meetings the ami and amida projects.
\newblock In \emph{ASRU}, pages 238--247. IEEE.

\bibitem[{Ritter et~al.(2010)Ritter, Cherry, and
  Dolan}]{DBLP:conf/naacl/RitterCD10}
Alan Ritter, Colin Cherry, and Bill Dolan. 2010.
\newblock \href {https://www.aclweb.org/anthology/N10-1020/} {Unsupervised
  modeling of twitter conversations}.
\newblock In \emph{HLT}.

\bibitem[{Sankar and Ravi(2019)}]{DBLP:journals/corr/abs-1907-02848}
Chinnadhurai Sankar and Sujith Ravi. 2019.
\newblock \href {http://arxiv.org/abs/1907.02848} {Deep reinforcement learning
  for modeling chit-chat dialog with discrete attributes}.
\newblock \emph{CoRR}, abs/1907.02848.

\bibitem[{See et~al.(2019)See, Roller, Kiela, and
  Weston}]{DBLP:conf/naacl/SeeRKW19}
Abigail See, Stephen Roller, Douwe Kiela, and Jason Weston. 2019.
\newblock \href {https://www.aclweb.org/anthology/N19-1170/} {What makes a good
  conversation? how controllable attributes affect human judgments}.
\newblock In \emph{NAACL-HLT}.

\bibitem[{Serban et~al.(2018)Serban, Lowe, Henderson, Charlin, and
  Pineau}]{DBLP:journals/dad/SerbanLHCP18}
Iulian~Vlad Serban, Ryan Lowe, Peter Henderson, Laurent Charlin, and Joelle
  Pineau. 2018.
\newblock \href {http://dad.uni-bielefeld.de/index.php/dad/article/view/3690}
  {A survey of available corpora for building data-driven dialogue systems: The
  journal version}.
\newblock \emph{D{\&}D}, 9(1).

\bibitem[{Serban et~al.(2016)Serban, Sordoni, Bengio, Courville, and
  Pineau}]{DBLP:conf/aaai/SerbanSBCP16}
Iulian~Vlad Serban, Alessandro Sordoni, Yoshua Bengio, Aaron~C. Courville, and
  Joelle Pineau. 2016.
\newblock \href
  {http://www.aaai.org/ocs/index.php/AAAI/AAAI16/paper/view/11957} {Building
  end-to-end dialogue systems using generative hierarchical neural network
  models}.
\newblock In \emph{AAAI}.

\bibitem[{Shang et~al.(2015)Shang, Lu, and Li}]{DBLP:conf/acl/ShangLL15}
Lifeng Shang, Zhengdong Lu, and Hang Li. 2015.
\newblock \href {https://www.aclweb.org/anthology/P15-1152/} {Neural responding
  machine for short-text conversation}.
\newblock In \emph{ACL}.

\bibitem[{Sordoni et~al.(2015{\natexlab{a}})Sordoni, Bengio, Vahabi, Lioma,
  Simonsen, and Nie}]{DBLP:conf/cikm/SordoniBVLSN15}
Alessandro Sordoni, Yoshua Bengio, Hossein Vahabi, Christina Lioma, Jakob~Grue
  Simonsen, and Jian{-}Yun Nie. 2015{\natexlab{a}}.
\newblock \href {https://doi.org/10.1145/2806416.2806493} {A hierarchical
  recurrent encoder-decoder for generative context-aware query suggestion}.
\newblock In \emph{CIKM}.

\bibitem[{Sordoni et~al.(2015{\natexlab{b}})Sordoni, Galley, Auli, Brockett,
  Ji, Mitchell, Nie, Gao, and Dolan}]{DBLP:conf/naacl/SordoniGABJMNGD15}
Alessandro Sordoni, Michel Galley, Michael Auli, Chris Brockett, Yangfeng Ji,
  Margaret Mitchell, Jian{-}Yun Nie, Jianfeng Gao, and Bill Dolan.
  2015{\natexlab{b}}.
\newblock \href {https://www.aclweb.org/anthology/N15-1020/} {A neural network
  approach to context-sensitive generation of conversational responses}.
\newblock In \emph{NAACL-HLT}.

\bibitem[{Strassel(2004)}]{darpa_ears}
Stephanie~M Strassel. 2004.
\newblock Linguistic resources for effective, affordable, reusable
  speech-to-text.
\newblock In \emph{LREC}.

\bibitem[{Weizman(2008)}]{weizman2008positioning}
Elda Weizman. 2008.
\newblock \emph{Positioning in media dialogue: Negotiating roles in the news
  interview}, volume~3.
\newblock John Benjamins Publishing.

\bibitem[{Zhang et~al.(2018)Zhang, Dinan, Urbanek, Szlam, Kiela, and
  Weston}]{DBLP:conf/acl/KielaWZDUS18}
Saizheng Zhang, Emily Dinan, Jack Urbanek, Arthur Szlam, Douwe Kiela, and Jason
  Weston. 2018.
\newblock \href {https://doi.org/10.18653/v1/P18-1205} {Personalizing dialogue
  agents: {I} have a dog, do you have pets too?}
\newblock In \emph{ACL}.

\bibitem[{Zhang et~al.(2019)Zhang, Sun, Galley, Chen, Brockett, Gao, Gao, Liu,
  and Dolan}]{DBLP:journals/corr/abs-1911-00536}
Yizhe Zhang, Siqi Sun, Michel Galley, Yen{-}Chun Chen, Chris Brockett, Xiang
  Gao, Jianfeng Gao, Jingjing Liu, and Bill Dolan. 2019.
\newblock \href {http://arxiv.org/abs/1911.00536} {Dialogpt: Large-scale
  generative pre-training for conversational response generation}.
\newblock \emph{CoRR}, abs/1911.00536.

\end{thebibliography}
\bibliographystyle{acl_natbib}
\appendix

\begin{table*}
[h!]
\small
\centering
\begin{tabularx}{\linewidth}{ p{19mm} X }
\toprule

\bf Context &\textbf{SCOTT SIMON}:  Airstrikes by Syria and the Russian allies have been exacting a high cost in civilian lives and could trigger yet another wave of refugees if they can find a place to go.

\textbf{SCOTT SIMON}:  Panos Moumtzis is the U.N.'s humanitarian coordinator for Syria and joins us now. Mr. Moumtzis, thanks so much for being with us.

\textbf{GUEST}:  Thank you for calling me.

\textbf{SCOTT SIMON}:  What do you hear from people on the ground in and around Idlib?

\textbf{GUEST}:  People on the ground are really panicking. Women, children, families have been, on a daily basis for the last four months, been in an area where there's been bombing, airstrikes, ground offensives that has really made their life impossible. More than 400,000 had to run to safety within the governorate in an area where 3 million people are living. We've had more than a hundred incidents of health clinics, schools, water distribution points, a market that received a direct strike, which was really - has tolled, so far, more than 500 deaths and a total panic where one sees entire villages and towns deserted, people running to safety.

\textbf{GUEST}:  So our biggest concern is really the protection of civilians. The basic rule of war - that residential areas, civilians should be respected and protected - it is violated at this point in the Idlib area. And that's really something that is not acceptable.

\textbf{SCOTT SIMON}:  We should remind people, in fact, a direct intentional attack on civilians is a war crime, isn't it?

\textbf{GUEST}:  It is. If it's proven that it's intentional, it is.

\textbf{GUEST}:  And, of course, we have humanitarian assistance programs on the ground providing lifesaving assistance to the thousands of people who are escaping, going into safety. The worry is, first of all, within Idlib itself, it's a governorate with 3 million people living. This is the area where many people had fled from throughout the country. So there's a displacement that is happening on the doorstep of Turkey with 3 million people inside. And it has, also, all the elements to create further regional destabilization should this escalation continue.

\textbf{GUEST}:  What we are calling for is a cessation of hostilities. There must be a peaceful way forward, a way which - a solution that does not involve residential areas being bombed, a solution that is not military and that - a solution that ensures the protection of the children, the women, the families, the civilians who are living in this area.

\textbf{SCOTT SIMON}: \\ \midrule
\bf Gold &  And there have also been strikes from the rebel side. \\ \midrule
\bf FT DialoGPT &  Mr. Moumtzis, I want you to tell me about some of the people who are in these areas that you're talking to. I know that \textcolor{red}{you're a refugee yourself, but you've also been involved in fighting in the north and the south of Syria.} How are they dealing with these attacks in Syria? \\ \midrule
\bf FT GPT2 &  What do you see in the future for the U.N.? \\ \midrule
\bf Speaker GPT2 &  Well, that's what you've been talking about, right? \\ \midrule
\bf Speaker DialoGPT &  Mr. Moumtzis, we should note that this is a very small group that has, you know, a very small number of members. And we have seen reports that there is a large number of people who live in the area, and they have been killed or wounded in the fighting, and that \textbf{there are people who live in areas that are being targeted}. What do you see as the biggest threat, the biggest danger, of the people that live in \textbf{Idlib}? \\

\bottomrule
\end{tabularx}
\caption{Sample generated response on Syrian air strikes. \textbf{Bold} emphasizes specificity and topicality. \textcolor{red}{Red} denotes factually incorrect or inconsistent segments.}
\label{tab:sample_1}
\end{table*}

\begin{table*}
[h!]
\small
\centering
\begin{tabularx}{\linewidth}{ p{19mm} X }
\toprule

\bf Context &\textbf{GUEST}:  And it's unclear to me even at this point if he had the authority to launch that special operations team or whether someone outside that chain of command and Department of Defense or Southern Command, or pardon me, in Special Operations Command, you know, overruled that. And it's never become obvious to me what that chain of command was. And it seemed to me that would be one of the first questions the congressional people are asking - doing the inquiries should've been asking.

\textbf{NEAL CONAN}:  So in other words, had the ambassador, the head of mission, been alive, he would've been in control? But once he was dead, there was nobody in control.

\textbf{GUEST}:  Well, that's one issue, but the second issue is whether or not that special operations team was under the mission's control. If it wasn't, that decision was made in Washington. And then in that case, all right, who made that decision and why did they make that decision, and where did that come from based - above at the desk on Washington and so forth. And those questions have never been answered or is, as far as I can tell, in the current congressional testimony, haven't even been asked. They seem to be dancing around who's to blame rather than to try to figure out exactly what happened. That really is - that's disturbing if we're going to try to do lessons learned on this thing.

\textbf{NEAL CONAN}:  The other agency that was involved and, speaking of dancing around, is the Central Intelligence Agency.

\textbf{GUEST}:  Well, the Central Intelligence Agency was and, you know, theoretically they also were under the control of the mission. But that's - again, and that's why the senior agency thing becomes so muddled, were they or were they not, and also that special operations team, I don't think, was a working for the CIA. I think they were, you know, DOD asset that obviously was operating in support of the mission there on the ground. So who was in-charge remains a question. Obviously, the deputy from his testimony didn't think he had the power to do that, nor was the special operations commander on the ground able to take independent action apparently without going back to D.C.

\textbf{NEAL CONAN}:  And...

\textbf{GUEST}:  So that really raises some questions that we really ought to get to before the next one of these things pops up.

\textbf{NEAL CONAN}: \\ \midrule
\bf Gold &  There was also confusion. Ordinarily there would have been military aircraft - Air Force planes on call on behalf of the - to support operations, if the mission called, but they were too far away. \\ \midrule
\bf FT DialoGPT &  Well, you know, the CIA, you're saying, has the agency been able to determine what went wrong in the past? Is it clear that that's been done? \\ \midrule
\bf FT GPT2 &  We've heard a lot of talk in recent weeks about the importance of a strong national security strategy. What's that strategy? Is there a sense that the United States is at risk if we don't get it? \\ \midrule
\bf Speaker GPT2 &  Well, that's NPR senior correspondent James Bamford speaking with us from New York. \\ \midrule
\bf Speaker DialoGPT &  But the question that's being asked now is: Is there a way in which the CIA is going to be able to do that \textbf{without the special counsel or the FBI or the president}, and \textbf{without the president's approval or the president's permission?} \\

\bottomrule
\end{tabularx}
\caption{Sample generated response. \textbf{Bold} emphasizes specificity and topicality.}
\label{tab:sample_2}
\end{table*}

\begin{table*}
[h!]
\small
\centering
\begin{tabularx}{\linewidth}{ p{19mm} X }
\toprule

\bf Context &\textbf{SCOTT SIMON}:  You could find Jim Dunne slinking around Death Valley in peak heat, perched on a tree limb in Milford, Mich., or shooting through chain-link fences and barbed wire in Arizona. Jim Dunne is credited with maybe inventing the art and business of automotive spy photography. He died this week at the age of 87.

\textbf{SCOTT SIMON}:  Michelle Krebs, his friend and director of automotive relations for AutoTrader.com, joins us from studios of WDET in Detroit. Thanks so much for being with us.

\textbf{GUEST}:  Oh, well, thank you for letting me talk about Jim.

\textbf{SCOTT SIMON}:  Well, how did he get into this line of business - spy photography in the auto industry?

\textbf{GUEST}:  Well, like many of us, it was a lot of things that came together. He was in reconnaissance in the Army in the '50s. He got his degree in industrial engineering here at Wayne State University. And then he had seven kids he had to put through college.

\textbf{GUEST}:  And so he invented this thing called spy photography. He realized that he needed a competitive edge against different magazines. He was at Popular Science and then Popular Mechanics. And he thought maybe taking pictures of these cars that the automakers don't want me to take pictures of would be a business.

\textbf{SCOTT SIMON}:  What are some of the things he'd do to gain access, as they say these days?

\textbf{GUEST}:  Well, one of the amazing ones was that he would dress up with a short-sleeved white shirt, a skinny black tie, a pocket protector and a (laughter) - and a clipboard and just march into the Chrysler plant, where he knew that they had secret models hidden inside - out in the back part of the plant. And he could take pictures of them there.

\textbf{SCOTT SIMON}: \\ \midrule
\bf Gold &  Oh, my gosh. And people fell for that 'cause he was so convincing. \\ \midrule
\bf FT DialoGPT &  And how did this man become an icon? \\ \midrule
\bf FT GPT2 &  And he did that. \\ \midrule
\bf Speaker GPT2 &  So he would have to be in a car dealership to get ahold of that stuff, right? I mean, \textbf{what would he have to show me?} \\ \midrule
\bf Speaker DialoGPT &  And what was it like to be a \textbf{spy photographer} at the time? \\

\bottomrule
\end{tabularx}
\caption{Sample generated response on auto spy photography. \textbf{Bold} emphasizes specificity and topicality.}
\label{tab:sample_3}
\end{table*}

\begin{table*}
[h!]
\small
\centering
\begin{tabularx}{\linewidth}{ p{19mm} X }
\toprule

\bf Context &\textbf{GUEST}:  We have a plant in Kentucky and Taiwan and Japan. And yes, we make large sheets of this glass, and then we put it into crates, and then we ship it to companies called finishers, most of them are in China, and those are the companies that actually cut the glass into the pieces, drill holes, polish, et cetera, into the shapes of the parts that go onto the devices.

\textbf{IRA FLATOW}:  Why can't you - I'm sorry, why can't you make it the shape itself beforehand?

\textbf{GUEST}:  Well, it's much cheaper, more effective, to make these large sheets of glass and then cut it later, as opposed to trying to mold - or mold from molten glass into a small part.

\textbf{IRA FLATOW}:  It's not having to do with maybe the glass being too tough or brittle to work with at that point?

\textbf{GUEST}:  No, that's a good question, but we - the chemical strengthening process actually happens after our immediate customers have cut the parts down to the size, the shape of the parts.

\textbf{IRA FLATOW}:  There's a famous story in Walter Isaacson's biography of Steve Jobs that says he actually came down at one point and demanded that Corning make him some glass for his - any truth to that?

\textbf{GUEST}:  Well, the story, we certainly, we don't comment in detail about the relationship with different customers, but the story is there in the book, and it's true.

\textbf{IRA FLATOW}:  Is there other kinds of products, now that you've discovered how popular and ubiquitous the Gorilla Glass could be - any other places we might see it showing up?

\textbf{GUEST}:  It's a really exciting time for us, Ira, to - it's only five years in, and it's been a tremendous success story for the company. Handsets really have taken off. We're at a point now where almost half of all handsets in the world made are smartphones with some type of touch capability.

\textbf{GUEST}:  And then I'm sure you know of tablet computers, are very exciting, have grown quite magnificently in the last few years. And the market is starting to think about expanding touch interfaces with the different devices. If you think about your laptop computer, for example, or monitors or tabletops, even the front of a refrigerator, there's a lot more applications out there, we feel.

\textbf{GUEST}:  We feel that touch and the interface of touch is only going to grow.

\textbf{IRA FLATOW}: \\ \midrule
\bf Gold &  Do you have to design and engineer the glass specifically to be touched as a utility? \\ \midrule
\bf FT DialoGPT &  I have to ask you, you know, I think \textcolor{red}{you mentioned that you have been working on this for a very long time}. \\ \midrule
\bf FT GPT2 &  I'm thinking of the iPad, for example. \\ \midrule
\bf Speaker GPT2 &  So if the \textbf{Gorilla Glass has been in the news}, I mean, you have a lot of people in the world looking at it and saying, well, \textbf{we want to make it thinner. But we don't have the power to make it thinner.} \\ \midrule
\bf Speaker DialoGPT &  You're saying that \textbf{the market is growing?} \\

\bottomrule
\end{tabularx}
\caption{Sample generated response. \textbf{Bold} emphasizes specificity and topicality.  \textcolor{red}{Red} denotes factually incorrect or inconsistent segments.}
\label{tab:sample_4}
\end{table*}

% a hidden size $d_h$ of 256 for both the encoder and decoder.
% The embedding dimensions for vocabulary, ingredient, recipe, techniques, and caloric level are 300, 10, 50, 50, and 5 (respectively).
% The dropout rate is set at 0.2.
% For prior recipe attentions, we choose $k$ as 20, the 80th \%-ile for number of recipes interacted by an user.
% We use the Adam optimizer \cite{DBLP:journals/corr/KingmaB14} with a learning rate $10^{-3}$.
% We trained with batch size 24-36 (depending on models) on a single NVIDIA 1080Ti GPUs, until convergence.
% We used learning rate annealing with decay rate 0.9 \cite{DBLP:conf/acl/RuderH18}.
% We also use teacher-forcing \cite{DBLP:journals/neco/WilliamsZ89} for all our training epochs. 

\section{Generated Examples}
See the following tables for sample dialog histories and generated host responses from each of our baseline and speaker-conditioned dialog models.

\end{document}